\documentclass{article}
\usepackage{spconf,amsmath,graphicx,hyperref}
\usepackage[ruled,vlined]{algorithm2e}
\usepackage{booktabs}
\usepackage{float}
\usepackage{amsmath,amssymb}
\SetKw{KwReturn}{Return}
\SetKwInput{KwInput}{Input}
\usepackage{caption}

\title{discrete-time diffusion-like models for speech synthesis}
\name{Xiaozhou Tan, \qquad Minghui Zhao, \qquad Anton Ragni \thanks{Thanks to Mattias Cross from University of Sheffield for answering some theoretical questions.}}

\address{Department of Computer Science, University of Sheffield, UK\\
\{xtan30, mzhao39, a.ragni\}@sheffield.ac.uk}

\begin{document}
\ninept
\maketitle
\begin{abstract}
Diffusion models have attracted a lot of attention in recent years. 
These models view speech generation as a continuous-time process. For efficient training, this process is typically restricted to additive Gaussian noising, which is limiting. For inference, the time is typically discretized, leading to the mismatch between continuous training and discrete sampling conditions. Recently proposed discrete-time  processes, on the other hand, usually do not have these limitations, may require substantially fewer inference steps, and are fully consistent between training/inference conditions. This paper explores some diffusion-like discrete-time processes and proposes some new variants. These include processes applying additive Gaussian noise, multiplicative Gaussian noise, blurring noise and a mixture of blurring and Gaussian noises. The experimental results suggest that discrete-time processes offer comparable subjective and objective speech quality to their widely popular continuous counterpart, with more efficient and consistent training and inference schemas. 
\end{abstract}
\section{Introduction}
\label{sec:intro}

Diffusion and diffusion-like models have recently garnered significant attention in the areas of image \cite{ho2020denoising}, language \cite{li2022diffusionlm}, and speech generation \cite{kong2021diffwave}. A diffusion model consists of both a noising and a denoising process, forming a data trajectory between clean data and fully corrupted, noisy, data. One of the most popular diffusion models (DM) is the score-based generative model \cite{song2021scorebased}, where scores are derivatives of log-likelihoods of noised data in a continuous-time $t$ space. These scores are at the core of stochastic differential equations (SDEs) that describe how to denoise noised data into clean data \cite{ho2020denoising}. 
Another popular diffusion-like model is the flow matching model (FM) \cite{lipman2022flow}. The core idea behind FM is to learn a velocity field that describes a probability path between a source distribution and a target distribution. This probability path defines the data trajectory over a continuous-time, and the velocity field corresponds to the time derivative of this trajectory.  Flow matching \cite{lipman2024flowmatchingguidecode}  generalizes diffusion models by interpreting the data trajectories generated through the noising process --- modeled by SDEs in diffusion models --- as probability paths between source and target distributions. These two models are similar in that they both perform a noising process in a continuous-time space in training and generate samples using discretized differential equation solvers  \cite{lipman2022flow}. During inference, these models start from a fully corrupted data/initial sample and iteratively refine it to obtain a clean sample.   

Despite their successful application to text-to-speech (TTS) \cite{Popov2021GradTTS,mehta2024matchattsfastttsarchitecture}, the majority of these works \cite{song2021scorebased,mehta2024matchattsfastttsarchitecture} only choose additive Gaussian noise as a noising method. This choice implicitly assumes that the noised data are locally well-modelled by isotropic Gaussians, which have covariance matrices proportional to the identity matrix. However, the covariance matrix of the data, such as the Mel spectrogram, is non-isotropic and depends on the underlying signal energy. Unfortunately, a more complex noising process can render core components --- scores (in DM) and velocity fields (in FM) ---analytically unsolvable, which is limiting, thus making training intractable.  In addition, DM and FM training and inference methods are inconsistent. Training is performed in a continuous-time space, whereas inference is performed in a discrete-time space as continuous-time inference would require an infinite number of inference steps. Another discrepancy between training and inference in score-based generative models \cite{song2021scorebased} is that the initial sample for inference is assumed to be drawn from the forward noising process at a time step approaching infinity, but during training, the model is never exposed to these extreme time steps. 
 
In contrast, diffusion-like models that have a fully discrete training and inference process have not been evaluated for TTS. Like DMs and FMs, they have an iterative refinement process during inference. Unlike DMs and FMs, they preserve the consistency between training and inference by performing both processes in discrete-time space $n$ and by using exactly the data exposed in training as the initial sample during inference. In addition, more diverse noise types, for example, multiplicative Gaussian noise, can be explored with these models. Unlike in DM or FM cases, no substantial changes are needed. The application of such discrete-time models in speech synthesis has lagged behind that in image generation, where blurring   \cite{rissanen2023generativemodellinginverseheat}, snowification   \cite{bansal2022colddiffusioninvertingarbitrary}, and noise mixture   \cite{hsueh2025warm} have been explored. This paper aims to address this gap by exploring discrete-time diffusion-like models with the following noising methods:

    \begin{itemize}
        \item Additive Gaussian noise. This allows comparison between fully discrete and partially discretized processes.
        \item  Multiplicative Gaussian noise. While it is commonly used to model signal-dependent variability in natural systems \cite{bioucas2009multiplicative,oksendal2003sde}, it is underexplored in DM and FM. Compared with commonly used additive Gaussian noise, it has a non-isotropic covariance matrix. Because real-world data's covariance matrix is highly likely non-uniform, multiplicative Gaussian noise might generalize better for real-world data, such as speech.   
        \item Blurring noise is an example of fully deterministic noise. Using blurring as a noising process examines the need for any randomness in diffusion-like models in speech synthesis.
        \item A mixture of blurring and Gaussian noise. This mixture of noise is explored for leveraging  the structured dependencies of the deterministic noising process and stochastic variability in the stochastic process \cite{hsueh2025warm}.
    \end{itemize}

The rest of the paper is organized as follows: Section \ref{sec:Preliminaries} presents the preliminaries of this work, and Section \ref{sec:Method} describes different noising processes, training, and inference methods explored in this work.
Section \ref{sec:Experiments} presents and discusses the experimental results. Section \ref{sec:Conclusion} provides the conclusion. 

\section{Preliminaries}
\label{sec:Preliminaries}
\subsection{Continuous-time diffusion-like models}


In continuous-time DMs \cite{song2021scorebased}, noised data at each time step can be obtained in closed form without simulation. Clean data  can be recovered by integrating the score function over a discretized stochastic differential equation (SDE) \cite{ho2020denoising}, which also has a corresponding non-stochastic formulation as an ordinary differential equation (ODE). Models can be trained to predict score  \cite{song2021scorebased}, noise  \cite{kingma2021variational}, or a mixture of clean data and noise  \cite{salimans2022progressive}. In all these models, scores are either directly predicted \cite{song2021scorebased} or computed \cite{kingma2021variational, salimans2022progressive}, and then used by discretized SDE or probability-flow ODE samplers for denoising.

FM learns a time dependent velocity field that describes a probability path between a source distribution and a target distribution. By viewing sample from source distribution as clean data and sample from target distribution as fully corrupted data, moving along the velocity field can be regarded as noising/denoising process. Inference in FM can be performed by numerically integrating the velocity field from the target (fully corrupted) distribution back to the source (clean) distribution. The common methods for defining the velocity field in flow matching model are optimal transport conditional flow matching (OT-CFM)    \cite{tong2024improving} and rectified flows  \cite{liu2022flowstraightfastlearning}. Both methods define a straight path between clean data $\mathbf{X}_0$ and fully corrupted data $\mathbf{X}_1$. In particular, a data trajectory between $\mathbf{X}_0$ and $\mathbf{X}_1$ in rectified flows is given by  
\begin{equation}
\mathbf{X}_t = (1 - t)\mathbf{X}_0 + t\mathbf{X}_1,
\label{eq:CRF}
\end{equation}
where $t$ belongs to $[0,1]$.  Both methods have also been successfully applied to speech synthesis  \cite{Guo2024VoiceFlow, mehta2024matchattsfastttsarchitecture,  prenger2018waveglowflowbasedgenerativenetwork,valle2020flowtronautoregressiveflowbasedgenerative}. Interestingly, although they predict a velocity field, the clean data can be computed through the deterministic link between velocity fields and clean data at any denoising step rather than just the terminal value $t=0$. For instance, in rectified flow \cite{liu2022flowstraightfastlearning}, $\mathbf{X}_0$ can be computed by simply subtracting the predicted velocity field at any time $t$ from $\mathbf{X}_1$. Thus any such noise prediction network could be viewed as a clean data prediction network rather than a noise prediction network as commonly described in the literature.

\vspace{-5pt}
\subsection{Discrete-time diffusion-like models}
\label{sec:discrete-timeDLM}

Interestingly, the popularity of diffusion models started from a discrete-time diffusion model --- denoising diffusion probabilistic model (DDPM)   \cite{ho2020denoising, sohldickstein2015deepunsupervisedlearningusing} rather than continuous.  DDPMs are Markov chains that progressively add Gaussian noise by transitioning from the start state, $n=1$, to the end state, $n=T$. The noise added at discrete-time index 
$n$ is predicted and used to compose an inverse Markov chain. Given noised data in state $n$, the inverse Markov chain is used to predict less noised data in state $n-1$. Although DDPMs could achieve high-fidelity results \cite{ho2020denoising}, sampling typically requires simulating a
Markov chain for many (thousands of) steps. While DDPM states are Markov, they generally do not need to be Markov. Non-Markovian noising and denoising processes have been shown to lead to more efficient models \cite{song2022denoisingdiffusionimplicitmodels}, capable of reducing the number of inference steps to 100s rather than 1000s. Deterministic Markov chains, implemented through deterministic noising processes such as blurring, have been explored in image processing \cite{bansal2022colddiffusioninvertingarbitrary}, but their application in speech synthesis remains largely underexplored.

\section{Method}
\label{sec:Method}
This section presents noising, training and inference methods used in this work. Four noising processes are explored for the first time in a discrete-time diffusion-like model in speech synthesis. 
\vspace{-8pt}
\subsection{Noising process}

In all noising processes, the corrupted data $\mathbf{X}_n$ in step $n$ can be calculated directly without simulation by 
\begin{equation}
\text{noising}(\mathbf{X}_0, \mathbf{U}, n) = \mathbf{X}_n, \quad n \in \{0, 1, 2, \ldots, N\},
\label{eq:noisingschema}
\end{equation}
where $\mathbf{X}_0$ is clean data, $\mathbf{U}=E_{\theta}(\mathbf{c})$ is the text embedding, where $\mathbf{c}$ is the input text. Although this restricts the range of possible noising methods, it provides an efficient training approach similar to the continuous-time methods of DM and FM.

\subsubsection{Additive Gaussian noise}
Two types of additive Gaussian noising processes are explored.

\textbf{In the first additive Gaussian noising process}, a discretized noising process adapted from a popular model Grad-TTS \cite{Popov2021GradTTS} is employed. 
\begin{align}
\mathbf{X}_n = &\left( I - e^{-\frac{1}{2}  \int_0^n \beta_s \, ds} \right) \mathbf{U} 
+ e^{-\frac{1}{2}  \int_0^n \beta_s \, ds} \mathbf{X}_0 \notag \\
&+ \int_0^n \sqrt{\beta_s} \, e^{\frac{1}{2}  \int_s^n \beta_u \, du} \, d\mathbf{W}_s,
\label{eq:DiscreGradTTS}
\end{align}
where $\mathbf{W}_t$ is a Brownian motion (BM), which is integrated over time and evaluated at discrete-time step $n$, $\beta_s$ is a linearly increasing function with respect to $s$. Discretizing the noising process in this model allows comparison between fully discrete and partially discretized processes. In the following, this adapted system is called `Grad-TTS-DT (discrete-time)’.

\textbf{In the second additive Gaussian noising process}, a straight path between clean data and corrupted data is explored
\begin{equation}
\mathbf{X}_n=  (1-\frac{n}{N})\mathbf{X}_0 + \frac{n}{N} (\epsilon_1 + \mathbf{U}),
\label{eq:RFAG}
\end{equation}
where $\epsilon_1 \sim \mathcal{N} (\mathbf{0},\sigma^2 \mathbf{I})$ and $\sigma$ is a hyperparameter. This system can be regarded as a discretized version of the rectified flow \eqref{eq:CRF} \cite{liu2022flowstraightfastlearning} or discrete-time flow matching. In a continuous noising process, straight paths are believed to be  preferred \cite{liu2022flowstraightfastlearning} because they correspond to deterministic couplings that do not increase transport cost under convex cost functions. This motivates exploring straight paths in a discrete-time space. A straight path is also applied in the following noising method. In the following, this system is called `RFAG (rectified flow with additive Gaussian noise)'.

\subsubsection{Multiplicative Gaussian noise}

In this case, additive Gaussian noise $\epsilon_1 + \mathbf{\mathbf{U}}$ in \eqref{eq:RFAG} is replaced with multiplicative noise $\epsilon_2 \cdot \mathbf{U}$, where $\epsilon_2 \sim \mathcal{N} (\mathbf{I},\sigma^2 \mathbf{I})$, and 
\begin{equation}
\mathbf{X}_n = (1-\frac{n}{N})\mathbf{X}_0 + \frac{n}{N} (\epsilon_2  \cdot \mathbf{ U}).
\label{eq:RFMG}
\end{equation}
The denoising process in this case starts with a multiplication between Gaussian noise and the prior. Multiplicative Gaussian noise is commonly used to model signal-dependent variability in natural systems, such as biological vision \cite{jeon2014developmental}, auditory perception \cite{angeloni2023dynamics}, and sensor imaging \cite{riutort2020bayesian}. Compared with linear diffusion paths of additive Gaussian noising process, multiplicative Gaussian introduces non-linear distortions. This leads to richer and more varied transformations potentially leading to more diverse and expressive outputs. In the following, this system is called `RFMG (rectified flow with multiplicative Gaussian noise)'.

\subsubsection{Blurring noise}
The blurring process is performed by applying a heat equation  \cite{rissanen2023generativemodellinginverseheat}
\begin{equation}
\text{blurring}(\mathbf{X}_0, n) 
= \mathbf{V} \exp(\mathbf{\Lambda} n) \mathbf{V}^\top \mathbf{X}_0, 
\label{eq:blurring}
\end{equation}
where $ n \in \{0, 1, 2, \ldots, N\}$,  $\mathbf{V}^\top$ is the cosine basis projection matrix \cite{rissanen2023generativemodellinginverseheat}. $\mathbf{\Lambda}$ is a negative matrix with the same shape $(W,H)$ as $\mathbf{X}_0$ and $\lambda_{i,j}$ given by $-\pi^2 \left( \frac{i^2}{W^2} + \frac{j^2}{H^2} \right)$,
where $i = 0, \ldots, W - 1$ and  $j = 0, \ldots, H - 1$. This operation smooths/averages out $\mathbf{X}_0$ in the noised samples
\begin{equation}
\mathbf{X}_n = (1-\frac{n}{N})\text{blurring}(\mathbf{X}_0,n) + \frac{n}{N} \mathbf{\mathbf{U}}. 
\label{eq:blurringnoising}
\end{equation}
This noising process is fully deterministic. Deterministic noising methods leverage structured dependencies between localized spectrotemporal features and global spectral patterns to enhance signal representation. In the following, this system is called 'Blurring'.

\subsubsection{Mixture of blurring
and Gaussian noises}
The mixture noise is drawn from a Gaussian distribution whose mean is the blurring noise  \eqref{eq:blurringnoising}
\begin{equation}
    \text{mixture\_noise}(\mathbf{X}_0, n) \sim \mathcal{N}(\text{blurring}(\mathbf{X}_0, n), \frac{1}{2}\mathbf{(-\Lambda)}  \odot \boldsymbol{{I}}),
    \label{eq:warm}
\end{equation}
where $\mathbf{\Lambda}$ is the same as the negative  matrix in the blurring process~\eqref{eq:blurring}. The constant value $\frac{1}{2}$ is chosen based on the best results achieved in a related image generation work \cite{hsueh2025warm}. This mixture noise is applied to $\mathbf{X}_0$ to yield noised samples as follows 
\begin{equation}
\mathbf{X}_n = (1-\frac{n}{N})\text{mixture\_noise}(\mathbf{X}_0,n) + \frac{n}{N} \mathbf{\mathbf{U}}. 
\label{eq:warmnoising}
\end{equation}
This noising process \cite{hsueh2025warm} mixes blurring  \eqref{eq:blurring} and Gaussian noise. Although blurring can exploit the structured dependencies, it disregards the role of noise (randomness) in structuring the data manifold \cite{Hoogeboom2022BlurringDiffusionModels}. 
Therefore, the above process takes advantage of the deterministic and stochastic noising processes by controlling blurring and noise jointly. In the following, this system is called `Mixture'.
\vspace{-5pt}
\subsection{Training}
\label{sec:Training}
The discrete-time diffusion-like models in this work are implementation following Grad-TTS \cite{Popov2021GradTTS}. In particular, the Monotonic Alignment Search (MAS) method followed by \cite{kim2020glowttsgenerativeflowtexttospeech} is used to train a duration predictor. The duration predictor is part of a text encoder which produces prior $\mathbf{U}=E_{\theta}(\mathbf{c})$. Using this prior, a discrete-time training process is applied to the decoder.

The decoder is a non-causal residual convolutional network $R_{\theta}$ adapted from Grad-TTS. The adaptation includes the removal of continuous-time embeddings and the prediction of clean data instead of scores. As is discussed in Section \ref{sec:Preliminaries}, the majority of diffusion-like models can be viewed as predicting the clean data. Also, predicting clean data directly is more consistent \cite{song2023consistencymodels} than predicting noise in different time steps and it generalizes across all noise levels. The training aims to minimize the mean square error (MSE) between predicted $\hat{\mathbf{X}}_0$  and clean data $\mathbf{X}_0$.
\begin{equation}
\mathcal{L}_{\text{clean\_data}} =
\mathrm{MSE}\!\left(
  R_{\theta}\!\big(F(\mathbf{X}_0, E_{\theta}(\mathbf{c}), n), E_{\theta}(\mathbf{c})\big),
  \mathbf{X}_0
\right).
\label{eq:training without n}
\end{equation}
The training alternates between alignment optimization (by finding alignment using MAS and optimizing $\mathcal{L}_{\text{enc}}$) and total loss optimization (by optimizing $\mathcal{L}_{\text{dur}}+\mathcal{L}_{\text{enc}}+\mathcal{L}_\text{clean\_data}$), where $\mathcal{L}_{\text{dur}}$ and $\mathcal{L}_{\text{enc}}$ are losses in the neural network $E_{\theta}(\mathbf{c})$ adopted from Glow-TTS \cite{kim2020glowttsgenerativeflowtexttospeech} and Grad-TTS \cite{song2021scorebased}.

\subsection{Inference}
\label{sec:Inference}
The inference process is presented in Algorithm~\ref{alg:FS} and is applied to all systems except the system with blurring noise whose inference process applying Algorithm~\ref{alg:SS} follows    \cite{bansal2022colddiffusioninvertingarbitrary}. Compared with Algorithm~\ref{alg:FS},  Algorithm~\ref{alg:SS} is better for smooth/differentiated noising because it corrects restoration errors using a first-order approximation of the smooth degradation process \cite{bansal2022colddiffusioninvertingarbitrary}. $\text{Corrupt}(E_{\theta}(\mathbf{c}),N)$ in Algorithm~\ref{alg:SS} is a function that provides fully corrupted data / initial sample to start inference. 


\begin{algorithm}
\caption{First Sampling Method}\label{alg:FS}
\KwInput{Input text $\mathbf{c}$}
$\mathbf{X}_N \leftarrow \text{Corrupt}(E_{\theta}(\mathbf{c}),N)$\;
\For{$n = N, N-1, \ldots, 1$}{
  $\mathbf{\hat{X}}_0 \leftarrow R_{\theta}(\mathbf{X}_n, E_{\theta}(\mathbf{c}))$\;
  $\mathbf{X}_{n-1} \leftarrow \text{noising}(\mathbf{\hat{X}}_0, E_{\theta}(\mathbf{c}), n-1)$\;
}
\KwReturn{$\mathbf{X}_0$}
\end{algorithm}
\vspace{-15pt}
\begin{algorithm}
\caption{Second Sampling Method}\label{alg:SS}
\KwInput{Input text $\mathbf{c}$}
$\mathbf{X}_N \leftarrow \text{Corrupt}(E_{\theta}(\mathbf{c}),N)$\;
\For{$n = N, N-1, \ldots, 1$}{
  $\mathbf{\hat{X}}_0 \leftarrow R_{\theta}(\mathbf{X}_n, E_{\theta}(\mathbf{c}))$\;
  $\begin{aligned}
  \mathbf{X}_{n-1} =\ & \mathbf{X}_n - \text{noising}(\mathbf{\hat{X}}_0, E_{\theta}(\mathbf{c}), n) \\
  & + \text{noising}(\mathbf{\hat{X}}_0, E_{\theta}(\mathbf{c}), n - 1)
  \end{aligned}$\;
}
\end{algorithm}
\vspace{-13pt}

\section{Experiments}
\label{sec:Experiments}
The dataset used in this work is LJspeech \cite{ljspeech17} which contains approximately 13,100 clips totaling 24 hours of American English female voice recordings sampled at 22.05kHz. We follow \cite{song2021scorebased} and split the data into training (around 12,000 clips), validation (around 100 clips), test (around 500 clips) set. 30 randomly selected texts from test set are used for Mean Opinion Score (MOS) evaluation.

Our work follows the pipeline of Grad-TTS, which includes an encoder $E_{\theta}(\mathbf{c})$, a decoder $R_{\theta}(., E_{\theta}(\mathbf{c}))$, and a fixed HIFI-GAN \cite{KongKB20} vocoder.
Each of our systems is trained by using an encoder $E_{\theta}(\mathbf{c})$ checkpoint from Grad-TTS, and training the decoder from scratch for 500 epochs with a frozen encoder. Then, the encoder is unfrozen and trained for another 100 epochs. The baseline model uses a checkpoint provided by Grad-TTS with either 5 or 10 inference steps \cite{Popov2021GradTTS}. Specifically, Grad-TTS is a score-based model which learns scores and integrates the score over a discretized SDE for inference. On the other hand, our implemented models are non-score based, where clean data is predicted  (Section~\ref{sec:Training}) and used to refine the result (Algorithm~\ref{alg:FS}, Algorithm ~\ref{alg:SS}) in each iterative step during inference (Section~\ref{sec:Inference}). The training and inference conditions are consistent in our models, meaning there is no discretization error.
\vspace{-8pt}
\subsection{Objective evaluation}

MCD, $\log f_0$, UTMOSv2 \cite{baba2024utmosv2} are used as objective metrics. The objective evaluation is performed on the whole test set. The evaluation result is presented in table \ref{tab:objectiveresult10steps}, where $\sigma$ represents the standard deviation of additive Gaussian noise \eqref{eq:RFAG} and multiplicative Gaussian noise \eqref{eq:RFMG} during noising process. The initial evaluation is conducted on results produced using 10 inference steps.
\vspace{-6pt}
\begin{table}[H]
\centering
\caption{Objective evaluation for the whole test set  (10 steps)}
\label{tab:objectiveresult10steps}  
\resizebox{0.48\textwidth}{!}{  
\begin{tabular}{lccc}
\toprule
\textbf{Systems} & \textbf{MCD} $\downarrow$ & \textbf{$\log f_0$}  $\downarrow$& \textbf{UTMOSv2} $\uparrow$\\
\midrule
Baseline & $5.71 \pm 0.46$ & $0.33 \pm 0.08$ & $4.03 \pm 0.22$ \\
Grad-TTS-DT& $5.53 \pm 0.52$ & $0.31 \pm 0.08$ & $3.95 \pm 0.21$ \\
\midrule
RFAG $(\sigma=0.2) $ & $5.55 \pm 0.55$ & $0.32 \pm 0.08$ & $3.96 \pm 0.23$ \\
RFAG $(\sigma=0.4)$ & $5.43 \pm 0.56$ & $0.32 \pm 0.09$ & $3.86 \pm 0.22$ \\
RFAG $(\sigma=0.6) $ & $5.49 \pm 0.52$ & $0.31 \pm 0.08$ & $3.86 \pm 0.22$ \\
RFMG $(\sigma=0.2)$ & $5.65 \pm 0.51$ & $0.32 \pm 0.07$ & $3.92 \pm 0.21$ \\
RFMG $(\sigma=0.4)$ & $5.63 \pm 0.50$ & $0.32 \pm 0.07$ & $3.94 \pm 0.23$ \\
RFMG $(\sigma=0.6)$ & $5.62 \pm 0.50$ & $0.32 \pm 0.08$ & $3.87 \pm 0.22$ \\
\midrule
Blurring & $5.52 \pm 0.54$ & $0.30 \pm 0.08$ & $3.71 \pm 0.25$ \\
Mixture & $5.58 \pm 0.50$ & $0.30 \pm 0.08$ & $3.87 \pm 0.22$ \\
\bottomrule
\end{tabular}
}
\end{table}

\vspace{-5pt}
 As is shown in table \ref{tab:objectiveresult10steps}, discretized baseline (Grad-TTS-DT) performs closely to the baseline, which shows fully discrete process can achieve comparable results to its partially discretized counterparts. Also, there is no significant difference between performance of DM style additive Gaussian noise (in Grad-TTS-DT)  and rectified flow style additive Gaussian noise (in RFAG) in discrete-time space.  Multiplicative Gaussian noise (in RFMG) performs similarly to additive Gaussian noise (in RFAG) in a discretized rectified flow data trajectory \eqref{eq:RFMG} \eqref{eq:RFAG}.  The Mixture system and Blurring system achieved the best $\log f_0$ result. The Mixture system's UTMOSv2 score is similar to RFAG's and RFMG's in certain $\sigma$ range, it is likely that Mixture system can perform better with different ratio between blurring and noise \eqref{eq:warm} \cite{hsueh2025warm}. Blurring's UTMOSv2 is the lowest, likely due to blurring smooths critical acoustic details without introducing sufficient randomness or variability, which limits its ability to improve generalization in speech naturalness prediction. 

\vspace{-3pt}
\begin{table}[h!]
\centering
\caption{Objective evaluation for the whole test set  (5 steps)}
\label{tab:results}
\begin{tabular}{lccc}
\toprule
\textbf{Systems} & \textbf{MCD} $\downarrow$& \textbf{$\log f_0$} $\downarrow$& \textbf{UTMOSv2} $\uparrow$\\
\midrule
Baseline                & $5.69 \pm 0.53$ & $0.34 \pm 0.08$ & $3.98 \pm 0.22$ \\
Grad-TTS-DT             & $5.52 \pm 0.54$ & $0.31 \pm 0.08$ & $3.93 \pm 0.22$ \\
\midrule
RFAG $(\sigma=0.2)$     & $5.57 \pm 0.53$ & $0.32 \pm 0.08$ & $4.01 \pm 0.20$ \\
RFAG $(\sigma=0.4)$     & $5.44 \pm 0.52$ & $0.33 \pm 0.09$ & $3.88 \pm 0.24$ \\
RFAG $(\sigma=0.6)$     & $5.51 \pm 0.53$ & $0.32 \pm 0.08$ & $3.84 \pm 0.23$ \\
RFMG $(\sigma=0.2)$     & $5.56 \pm 0.53$ & $0.33 \pm 0.08$ & $3.87 \pm 0.22$ \\
RFMG $(\sigma=0.4)$     & $5.55 \pm 0.52$ & $0.33 \pm 0.08$ & $3.91 \pm 0.24$ \\
RFMG $(\sigma=0.6)$     & $5.55 \pm 0.52$ & $0.33 \pm 0.08$ & $3.86 \pm 0.23$ \\
\midrule
Blurring                & $5.53 \pm 0.56$ & $0.31 \pm 0.08$ & $3.66 \pm 0.24$ \\
Mixture                 & $5.52 \pm 0.53$ & $0.31 \pm 0.08$ & $3.87 \pm 0.22$ \\
\bottomrule
\end{tabular}
\end{table}
Objective evaluation on results from 5 steps inference is performed to check how our systems generalize to fewer time steps. Our systems have better MCD and $\log f_0$. In the UTMOSv2 evaluation, all systems performed similarly to the baseline and shows good generalization, except the Blurring system which might lack some randomness.  
\vspace{-8pt}
\subsection{Subjective evaluation}

The subjective evaluation is carried out on the Amazon Mechanical Turk (AMT) platform. 30 randomly selected texts from test set are used throughout the subjective evaluation. Each system is evaluated on 30 distinct speech samples, each rated by 20 native speakers. The ground\_truth result is generated by the fixed HIFI-GAN vocoder.

\vspace{-5pt}
\begin{table}[H]
\centering
\captionsetup{skip=2pt}
\caption{Subjective scores for 30 randomly selected speech}
\begin{tabular}{lcc}
\hline
\textbf{Systems} & \textbf{MOS} $\uparrow$\\
\hline
Ground truth         & $4.15 $ \\
\midrule
Baseline             & $4.02 $ \\
Grad-TTS-DT & $4.07 $ \\
\midrule
RFAG $(\sigma = 0.6)$        & $3.90 $ \\
RFMG $(\sigma = 0.6)$        & $3.89 $ \\
Blurring              & $3.86 $ \\
Mixture              & $3.86 $ \\
\hline
\end{tabular}
\label{tab:naturalness_scores}
\end{table}

\vspace{-20pt}

\begin{figure}[H]
    \centering
    \captionsetup{skip=1pt} 
    \includegraphics[scale=0.37]{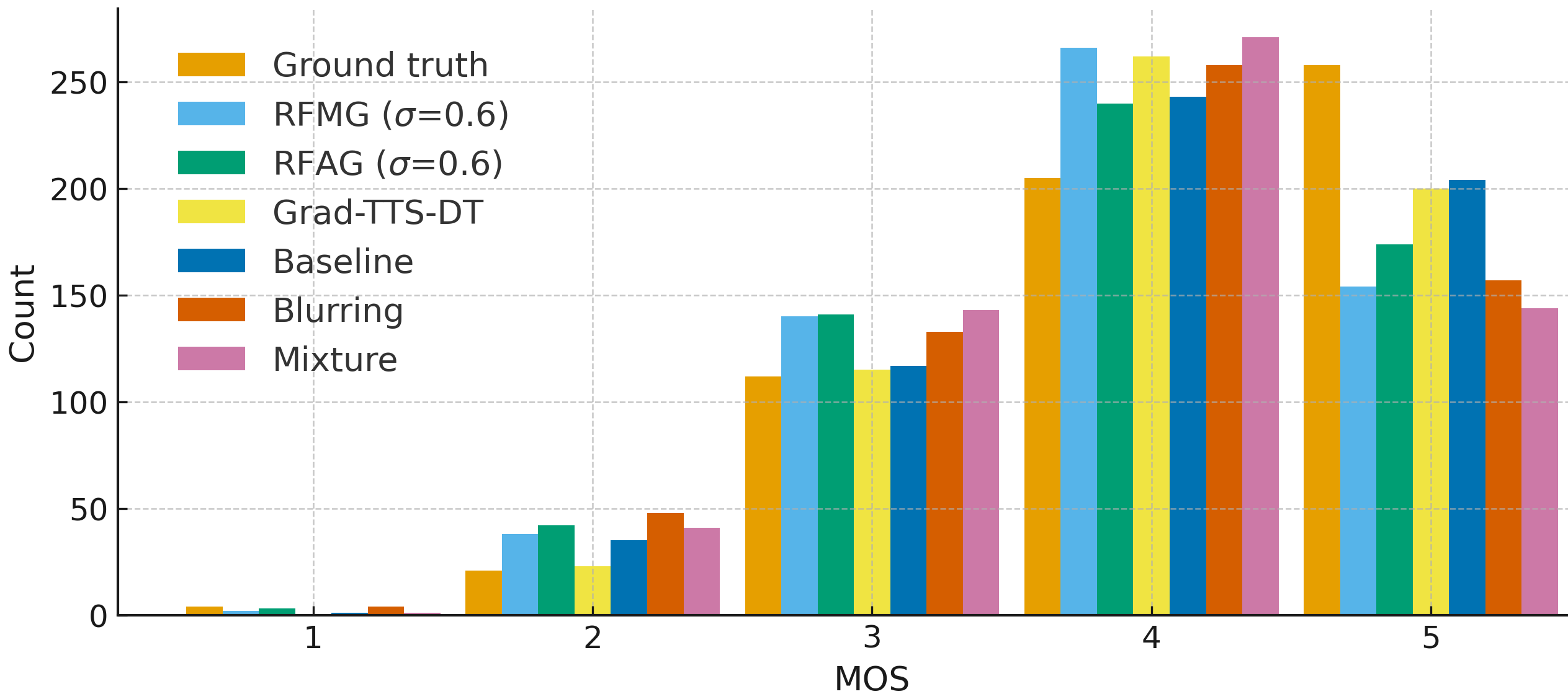}  

    \caption{Detailed breakdown of MOS score counts}
    \label{fig:example}
\end{figure}
\vspace{-8pt}

All our systems achieved similar MOS scores to the baseline.  It is also shown by Fig.~\ref{fig:example} that, aside from Ground truth system, the MOS results are distributed similarly across the different MOS levels. In MOS evaluation, Grad-TTS-DT slightly outperformed the baseline, demonstrating the effectiveness of a fully discrete process compared with the partially discretized process.
In addition, the encoder used in our systems is suboptimal and we believe that with tuning most of our systems could perform better.

\section{Conclusion}

\label{sec:Conclusion}
Due to the difficulty of implementing more complex noising processes in continuous-time diffusion-like models and the inconsistency between training and inference in these models, this work proposes discrete-time diffusion-like models. This work presented discrete-time diffusion-like models with four different noising processes. This is the first work to implement multiplicative Gaussian noise in a diffusion-like model, investigate blurring and a combination of blurring with Gaussian noise for speech synthesis using a diffusion-like model, and examine popular continuous-time diffusion-like models within a consistent, fully discrete-time framework. The results demonstrate that discrete-time diffusion models can perform comparably well to a popular continuous diffusion model. The performance of blurring suggests that randomness might still be needed for such discrete-time models. Additionally, noising methods such as multiplicative Gaussian noise, a mixture of blurring and additive Gaussian noise, can have a performance similar to that of the widely used additive Gaussian noise in speech generation tasks.

\clearpage
\vfill\pagebreak

\bibliographystyle{IEEEbib}
\bibliography{strings,refs}

\end{document}